\documentclass[conference]{IEEEtran}
\IEEEoverridecommandlockouts
\usepackage{cite}
\usepackage{amsmath,amssymb,amsfonts}
\usepackage{graphicx}
\usepackage{textcomp}
\usepackage[ruled, vlined, linesnumbered]{algorithm2e}
\usepackage[noend]{algorithmic}

\usepackage{booktabs} 
\usepackage{xcolor}         

\usepackage{bm}
\usepackage{subcaption}
\usepackage{bbding}
\usepackage{url}

\begin{document}

\title{A Simple and Scalable Graph Neural Network for Large Directed Graphs}

\author{\IEEEauthorblockN{Seiji Maekawa\thanks{This work was  done when Seiji Maekawa was a doctor student at Osaka University.}}
\IEEEauthorblockA{\textit{Megagon Labs}\\
Mountain View, USA \\
seiji@megagon.ai}
\and
\IEEEauthorblockN{Yuya Sasaki}
\IEEEauthorblockA{\textit{Osaka University}\\
Suita, Japan \\
sasaki@ist.osaka-u.ac.jp}
\and
\IEEEauthorblockN{Makoto Onizuka}
\IEEEauthorblockA{\textit{Osaka University}\\
Suita, Japan \\
onizuka@ist.osaka-u.ac.jp}
}

\newcommand{\bmC}{\bm{C}}
\newcommand{\bmH}{\bm{H}}
\newcommand{\bmX}{\bm{X}}
\newcommand{\bmS}{\bm{S}}
\newcommand{\bmD}{\bm{D}}
\newcommand{\bmA}{\bm{A}}
\newcommand{\bmI}{\bm{I}}
\newcommand{\bmW}{\bm{W}}
\newcommand{\bmY}{\bm{Y}}
\newcommand{\bmZ}{\bm{Z}}
\newcommand{\bmP}{\bm{P}}
\newcommand{\bmM}{\bm{M}}
\newcommand{\inmathbbr}[2]{\in \mathbb{R}^{#1 \times #2}}

\definecolor{mycolor}{RGB}{216, 216, 255}

\maketitle

\begin{abstract}
Node classification is one of the hottest tasks in graph analysis. 
Though existing studies have explored various node representations in directed and undirected graphs, they have overlooked the distinctions of their capabilities to capture the information of graphs.
To tackle the limitation, we investigate various combinations of node representations (aggregated features vs. adjacency lists) and edge direction awareness within an input graph (directed vs. undirected). 
We address the first empirical study to benchmark the performance of various GNNs that use either combination of node representations and edge direction awareness.
Our experiments demonstrate that no single combination stably achieves state-of-the-art results across datasets, which indicates that we need to select appropriate combinations depending on the dataset characteristics. 
In response, we propose a simple yet holistic classification method A2DUG which leverages all combinations of node representations in directed and undirected graphs. 
We demonstrate that A2DUG stably performs well on various datasets and improves the accuracy up to 11.29 compared with the state-of-the-art methods.
To spur the development of new methods, we publicly release our complete codebase under the MIT license. {\url{https://github.com/seijimaekawa/A2DUG}}
\end{abstract}

\section{Introduction}
\label{sec:intro}
The semi-supervised node classification task is one of the hottest topics in graph analysis. 
Its goal is to predict unknown labels of the nodes by using the topology structure and node attributes, given partially labeled networks.
Many algorithms \cite{kipf2017semi,monti2017geometric,hamilton2017inductive,velickovic2018graph,klicpera2018predict,xu2018powerful,zhu2020beyond,sign_icml_grl2020,zhu2021graph,shadow_GNN,maurya2021improving,chien2021adaptive,lim2021large,li2022finding} have been proposed to tackle this task and they have gained wide research interest from various domains including chemistry~\cite{duvenaud2015convolutional,fung2021benchmarking}, physics~\cite{sanchez2020learning}, social science~\cite{kipf2017semi}, and neuroscience~\cite{zhong2020eeg}. 

\subsection{Motivation}

To improve the classification quality, existing studies have explored various node representations in directed and undirected graphs.
However, they have overlooked the distinctions of the capabilities of the representations to capture the information of graphs. 
Our focus is on discussing the distinctions in terms of two key aspects: 1) node representations (aggregated features vs. adjacency lists), and 2) edge direction awareness within input graphs (directed vs. undirected), and then proposing a simple approach that leverages the benefits of each distinction.

\smallskip \noindent \textbf{Aggregated feature vs. Adjacency list. }
Most studies \cite{kipf2017semi,hamilton2017inductive} aim to construct node representations by incorporating two characteristics, node features and local topological structure.
There are two major approaches to constructing node representations to capture these characteristics: 1) Graph Neural Networks (GNNs) adopting feature aggregation~\cite{tong2020digraph,zhang2021magnet,kipf2017semi,velickovic2018graph,luan2022revisiting,chien2021adaptive} and 2) methods using adjacency lists as node representations~\cite{CDE2018,lim2021large,li2022finding}. 
The first approach captures node features by adopting
feature aggregation that aggregates node features from local neighbor nodes to obtain denoised features (we call \textit{aggregated features}).
In contrast, the second approach captures 1-hop topological information by utilizing an adjacency list as its node representation.
Therefore, these approaches capture different characteristics of node features and local topological structure. 
To facilitate understanding, we provide intuitive examples in Section \ref{ssec:motivating}.

\smallskip \noindent \textbf{Directed graph vs. Undirected graph. }
Most methods focus on undirected graphs \cite{kipf2017semi,klicpera2018predict,xu2018powerful,zhu2020beyond,zhu2021graph,shadow_GNN,maurya2021improving,chien2021adaptive}.
Since edges in most graphs are potentially associated with direction information that may contribute to classification quality, several existing studies have addressed a node classification task for directed graphs. 
First, a small number of GNN algorithms \cite{tong2020directed,tong2020digraph,zhang2021magnet} utilizes feature aggregation in directed graphs rather than in undirected graphs.  
Second, as for methods using adjacency lists, users can choose whether to use a directed or undirected adjacency list for input. 
However, the choice of utilizing a directed or undirected graph depends on the characteristics of datasets. We will give intuitive examples in Section \ref{ssec:motivating} for deeper understandings. 
This characteristic difference underscores the challenge users face in determining the appropriate method beforehand.

\begin{table*}[t]
    \caption{Node representations and edge direction awareness of existing methods and \textsc{A2DUG}.
    \CheckmarkBold (\XSolid) indicates that methods (do not) support the node representation and edge direction. 
    \CheckmarkBold$~^*$ indicates that methods use either a directed or undirected graph. 
    }
    \centering\scalebox{1}{
    \begin{tabular}{l|cccc} \toprule
        Methods & Aggregated feature & Adjacency list & Directed graph & Undirected graph \\\midrule 
        GNNs for undirected graphs & \CheckmarkBold & \XSolid & \XSolid& \CheckmarkBold\\
        GNNs for directed graphs & \CheckmarkBold & \XSolid & \CheckmarkBold & \CheckmarkBold \\
        Methods using adjacency lists & \XSolid & \CheckmarkBold & \CheckmarkBold$~^*$  & \CheckmarkBold$~^*$ \\ \midrule
        \textsc{A2DUG} (proposal) & \CheckmarkBold & \CheckmarkBold & \CheckmarkBold & \CheckmarkBold \\
        \bottomrule
    \end{tabular}
    }
\label{tb:02_comparison}
\end{table*}

\begin{figure*}[ht]
\centering
  \includegraphics[width=16cm]{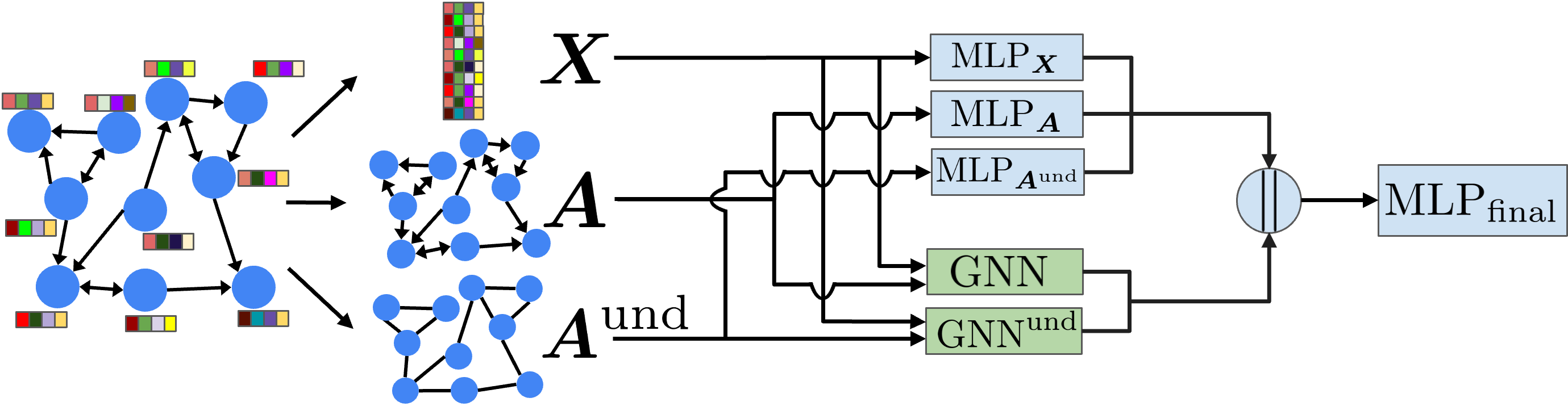}
  \caption{Our proposal \textbf{A2DUG} leverages both aggregated features and adjacency lists in directed/undirected graphs through GNN and MLP, respectively. 
  $\text{MLP}_{\bmX}, \text{MLP}_{\bmA}$, and $\text{MLP}_{\bmA^\text{und}}$ create node representations for node features $\bmX$, adjacency lists $\bmA$, and undirected adjacency lists $\bmA^\text{und}$. 
  GNN and $\text{GNN}^\text{und}$ compute aggregated features by inputting node features and adjacency lists.
  Finally, it concatenates the representations from MLP and GNN, and then outputs the results via $\text{MLP}_\text{final}$. 
  For the sake of brevity, we omit a transposed adjacency list in this figure (see the details in Section \ref{sec:proposal}). 
  }
  \label{fg:01_framework}
\end{figure*}

The above discussion reveals that each combination of node representations (aggregated features vs. adjacency lists) and edge direction awareness within input graphs (directed vs. undirected) holds unique distinguishability distinctions for the node classification task. 
This means that users face the challenge of selecting an appropriate method by considering whether to use aggregated features or adjacency lists and whether to treat the input graph as directed or undirected since existing methods focus on only one or two combinations as summarized in Table \ref{tb:02_comparison}.
However, it is burdensome and difficult to choose an appropriate method for each dataset particularly when users are not familiar with the dataset characteristics. 
To address this limitation and explore new possibilities, we aim to investigate the performance of methods utilizing different combinations of node representations and edge direction awareness across various directed graphs, which no comprehensive studies have explored. 
Furthermore, the difference in their unique distinguishability distinctions motivates us to design a new method that incorporates the advantages of all distinctions.

\subsection{Contributions}
First, we address the following research question: \textbf{Q1. } \textit{To what extent do node representations (aggregated features vs. adjacency lists) and edge direction awareness (directed vs. undirected) affect node classification quality?}
To answer this question, we conduct experiments using various existing methods including state-of-the-art methods on $13$ various directed graphs.
The experiments demonstrate that no existing method stably obtains the state-of-the-art classification results since the results highly depend on the characteristics of graphs, e.g., edge direction-aware methods perform well on a graph in which the prediction target is publication years (\texttt{arxiv-year} \cite{lim2021large}) and edge direction-unaware methods perform well on a graph in which the prediction target is research fields (\texttt{ogbn-arxiv} \cite{hu2020ogb}). 
Hence, we address the following research question: \textbf{Q2. } \textit{Can we adaptively choose the best combination of node representations and edge direction awareness for each dataset?}
To answer this question, we propose a simple yet holistic method, 
\textbf{A2DUG},
that leverages all combinations of node representations in directed and undirected graphs, as illustrated in Figure \ref{fg:01_framework}.
Further, it achieves high scalability by adopting precomputation-based GNNs \cite{wu2019simplifying,maurya2021improving}. 
Finally, we explore a potential model architecture extension and show the validity of A2DUG's model design.

We summarize our contributions as follows. 
\begin{itemize}
    \item {\bf Technical novelty}. We propose a simple yet scalable, effective, and robust method, A2DUG. It achieves higher accuracy than state-of-the-art methods in various datasets. In particular, it significantly increases the accuracy in middle/large-scale (from a million to a few hundred million edges) datasets, despite its simple architecture.
    \item {\bf Benchmarking}. We validate that no existing methods stably perform well across datasets. We show insufficiency to use only either combination of node representations and edge direction awareness
    \item {\bf Discovery}. We identify how the (un)usefulness of edge direction, aggregated feature, and adjacency list varies across different datasets. In a nutshell, undirected edges are effective in homophilous graphs, but there are no consistent trends in non-homophilous graphs.
\end{itemize}

We hope that this paper demonstrates the potential of future research on the combination of aggregated features and adjacency lists in directed/undirected graphs.

\subsection{Organization}
The rest of this paper is organized as follows. 
We describe the notation and problem definition in Section~\ref{sec:preliminaries}. We also position A2DUG with respect to the state of the art in the section. 
We propose A2DUG and discuss its complexity compared with existing methods in Section~\ref{sec:proposal}. 
Section \ref{sec:experiments} gives a detailed experimental analysis of the effectiveness and efficiency of A2DUG and existing methods. 
Section \ref{sec:discussion} summarizes the key insights that we obtain through the experiments. 
We give a conclusion and future work in Section~\ref{sec:conclusion}.

\section{Preliminaries}
\label{sec:preliminaries}

In this section, we describe notations that we used in this paper, problem definition, and motivating examples in our study.

\subsection{Notations and Problem Definition}
An \textit{attributed graph with class labels} is a triple $G=(\bmA,\bmX,\bmC)$ where $\bmA \in \{0,1\}^{n\times n}$ is an adjacency matrix, $\bmX\in \mathbb{R}^{n\times d}$ is an attribute matrix assigning attributes to nodes, 
and a class matrix $\bmC\in\{0,1\}^{n\times y}$ contains class information of each node,
and $n,d,y$ are the numbers of nodes, attributes, and classes, respectively. 
If there is a directed edge from node $i$ to node $j$,  $\bmA_{ij} = 1$ otherwise $\bmA_{ij} = 0$.
We denote a transposed adjacency matrix by $\bmA^{\top}$, where $\bmA^{\top}_{ij} = \bmA_{ji}$. 
Given an adjacency matrix $\bmA$, we can obtain an undirected adjacency matrix $\bmA^\text{und} \in \{0,1\}^{n\times n}$, where $\bmA^\text{und}_{ij} = \bmA^\text{und}_{ji} = 1$ if $\bmA_{ij} = 1$ or $\bmA_{ji} = 1$ otherwise $\bmA^\text{und}_{ij} = \bmA^\text{und}_{ji} = 0$. 
We define the degree matrix $\bmD=\text{diag}(D_1,\dots,D_n)\in\mathbb{R}^{n\times n}$ as a diagonal matrix, where $D_i$ expresses the degree of node $i$.
We also define an identity matrix $\bmI=\text{diag}(1,\dots,1)\in\mathbb{R}^{n\times n}$.
Table~\ref{tb: variables} lists the main symbols and their definitions for the following descriptions.
\begin{table}[t]
  \caption{Definition of main symbols.}
  \label{tb: variables}
  \begin{center}
	\begin{tabular}{l|l} \toprule
	Variable & Explanation \\ \midrule
    $n$ & number of nodes \\
    $d$ & number of attributes \\
    $y$ & number of classes \\
    $h$ & size of hidden dimension \\
    $k$ & number of layers \\
    $T$ & number of training epochs \\
    \midrule
    $\bmA \in\{0,1\}^{n\times n}$ &  adjacency matrix \\ 
    $\bmS \inmathbbr{n}{n}$ &  normalized adjacency matrix \\ 
    $\bmX \inmathbbr{n}{d}$ &  attribute matrix \\
    $\bmC \in \{1,\dots,k\}^{n}$ & class label \\ 
    $\bmY \inmathbbr{n}{y}$ & predicted class label \\ 
    $\bmD \inmathbbr{n}{n}$ &  degree matrix \\ 
    $\bmI \in\{0,1\}^{n\times n}$ &  identity matrix \\ 
    $\bmH \inmathbbr{n}{h}$ & node representation matrix \\ 
    \bottomrule
 	\end{tabular}
 \end{center}
\end{table}

We define two types of graphs, homophilous and non-homophilous graphs based on the edge homophily~\cite{zhu2020beyond}. 
The edge homophily is calculated as $H(G) = \frac{\sum_{0 \leq i,j < n} \bmA_{ij} \delta(\bmC_i,\bmC_j)}{\sum_{0 \leq i,j < n} \bmA_{ij}}$, where  $\delta(\bmC_i,\bmC_j)$ returns one if $\bmC_i = \bmC_j$ otherwise zero.
If $H(G)$ is large, the graph is homophilic, otherwise heterophilic.
Intuitively, in homophilous and non-homophilous graphs, nodes in the same class tend and do not tend to be connected, respectively.
It is well-known that GNNs for homophilic graphs do not work well in non-homophilic graphs.

In this paper, we focus on node classification tasks as follows:

\smallskip \noindent \textbf{Problem Definition (Node classification). } 
We split nodes into train/validation/test sets. 
Given an adjacency matrix $\bmA$, an attribute matrix $\bmX$, and a partial class matrix $\bmC'$ which contains class information of nodes in the train/validation sets, we predict the labels of the nodes in the test set.

\subsection{Motivating Examples}
\label{ssec:motivating}
In this subsection, we provide intuitive examples to explain the unique distinctions in term of two aspects: node representations and edge direction awareness. 

\begin{figure}
    \centering
    \includegraphics[width=8.7cm]{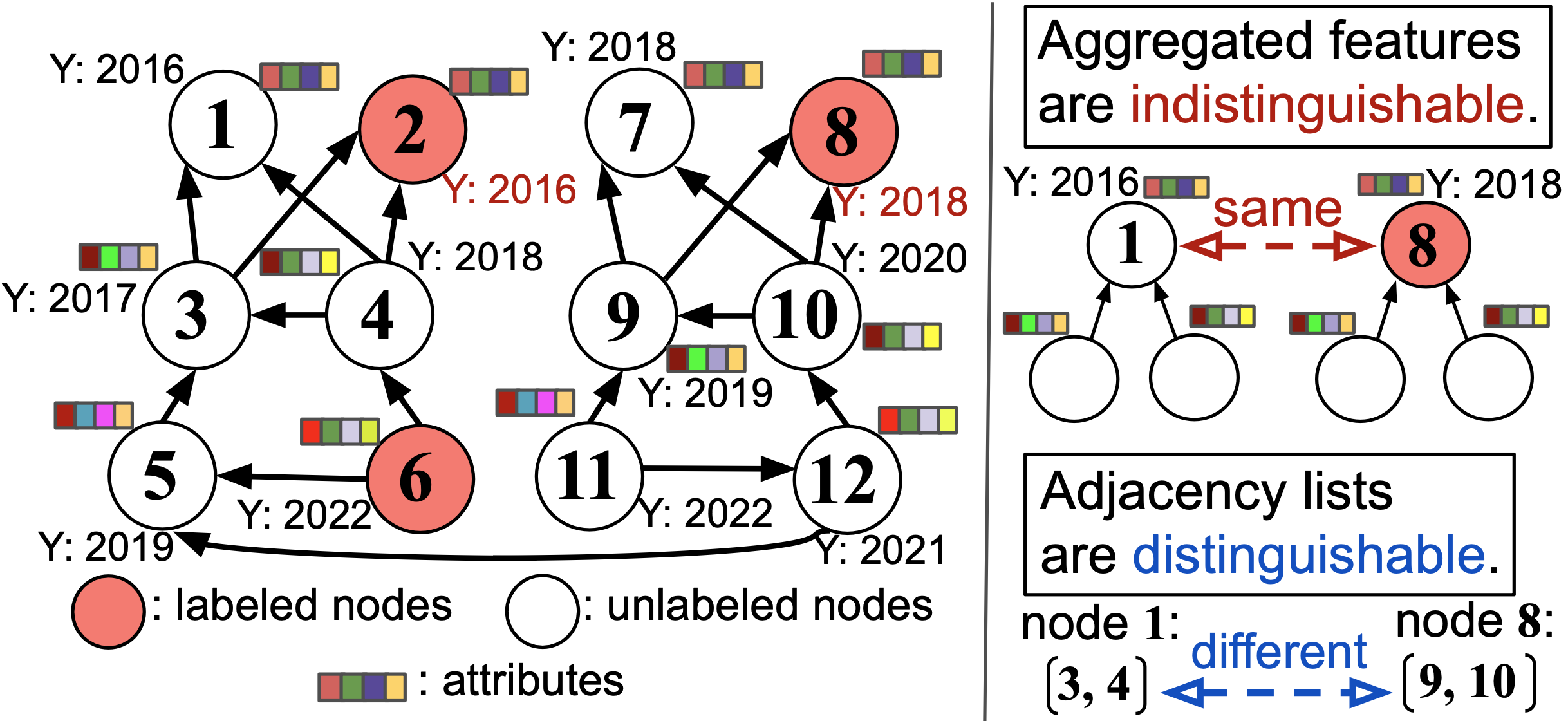}
    \caption{Example graph, in which labels indicate the publication years of papers. The red and white nodes indicate labeled and unlabeled nodes belonging to training and validation/test sets, respectively. 
    The aggregated features of nodes $1, 2, 7$, and $8$ are the same though nodes $1$/$2$ and nodes $7$/$8$ have different labels (2016 and 2018, respectively). 
    Thus, methods using aggregated features fail to predict the label of either node $1$ or $7$. 
    In contrast, methods using adjacency lists can distinguish the nodes and thus correctly predict them, e.g., nodes $1$ and $7$ are predicted as 2016 and 2018, respectively. 
    This is because their labels are the same as those of nodes $2$ and $8$ which have the same neighbors, respectively. 
    }
    \label{fg:01_year}
\end{figure}

\smallskip \noindent \textbf{Node representation (Aggregated feature vs. Adjacency list). }
First, we discuss the distinguishability distinctions between the two node representations by giving an intuitive example in Figure \ref{fg:01_year}, where the prediction target is the publication year of each paper.
On one hand, nodes $1$ and $8$, which are structurally distant\footnote{Their shortest path length is $5$ though they share the same node features in their neighbors.}, are not distinguishable by their aggregated features, because the aggregated features lose topologically different information 
and become the same representations, i.e., the same node features within 1-hop and 2-hop, resulting in identical representations (see the top in the right column of the figure).
On the other hand, using adjacency lists, nodes $1$ and $8$ are distinguishable as they connect to different adjacent nodes (see the bottom in the right column of the figure). 
Since nodes $1$ and $2$ share the same adjacency lists and node features, the label of node $1$ is correctly predicted as $2016$, the same as that of node $2$. 
Similarly, the label of node $7$ is accurately predicted as $2018$ using information from the labeled node $8$. 
This observation demonstrates that aggregated features lose local topological information, such as expressed by adjacency node IDs, even though feature aggregation utilizes adjacency lists in its calculation.

\begin{figure}
    \centering
    \includegraphics[width=8.6cm]{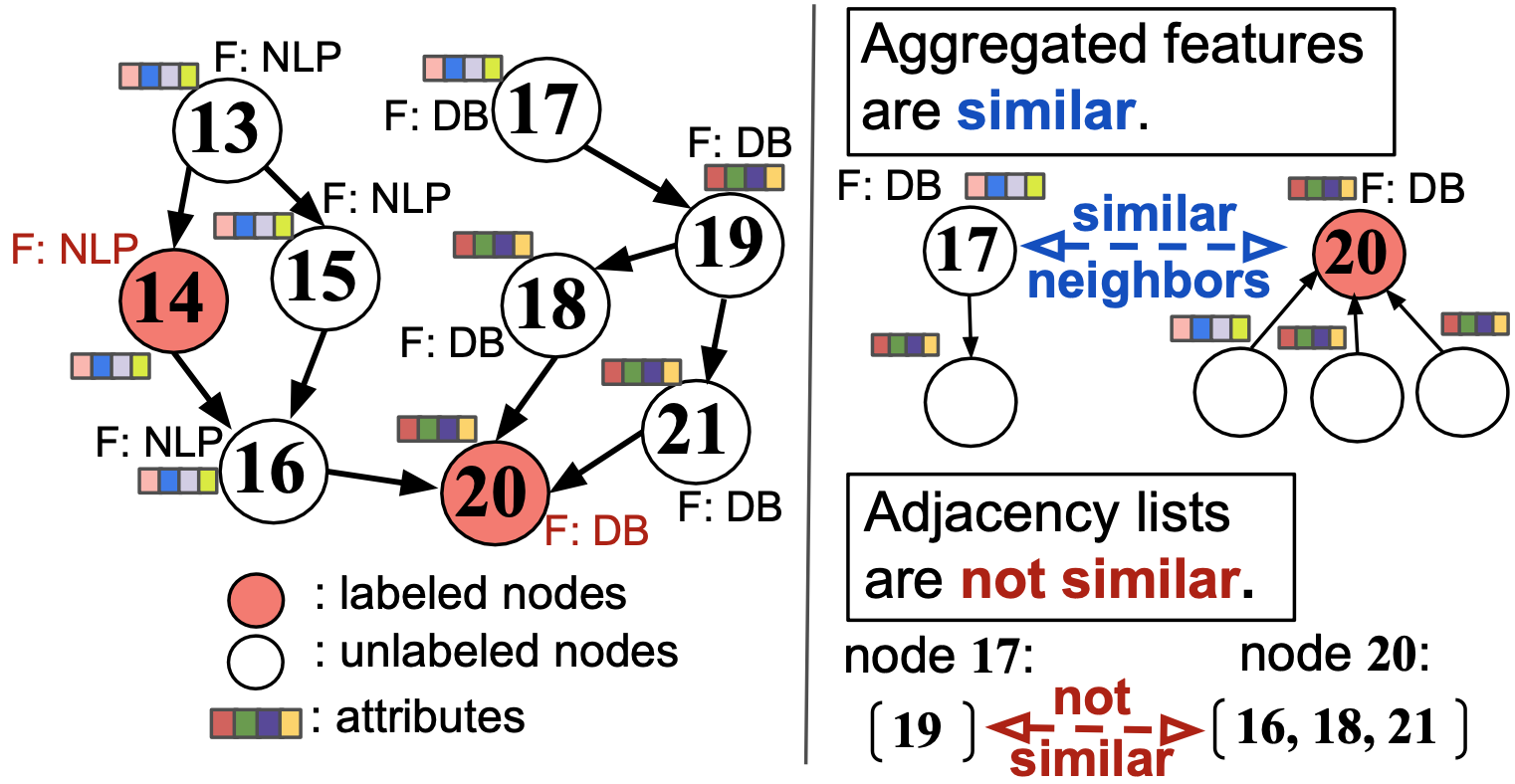}
    \caption{Second example: Labels indicate the research fields of papers. 
    Since the aggregated features of nodes $17$ and $20$ are similar, the methods using them can correctly predict the label of node $17$, i.e., DB. 
    In contrast, since the adjacency lists of nodes $17$ and $20$ are disjoint, the methods using adjacency lists fail to predict the label of node $17$. 
    }
    \label{fg:01_field}
\end{figure}
Next, Figure \ref{fg:01_field} shows another graph in which labels indicate the research fields of papers, e.g., NLP and DB. 
As for methods using aggregated features, node $17$ obtains a representation similar to that of a locally neighboring node $19$ via feature aggregation. 
Since the resulting representation is similar to that of the labeled node $20$, the methods are expected to correctly predict the label of node $17$, i.e., ``DB'' (see the top in the right column of the figure).
In contrast, the methods using adjacency lists fail to predict the label of node $17$ because there are no intersections between the adjacency lists of node $17$ and the labeled node $20$, which shares the same label as node $17$ (see the bottom in the right column of the figure and note that we do not consider edge direction in this case).

\smallskip \noindent \textbf{Edge direction awareness (Directed graph vs. Undirected graph). }
We use the former two examples to show that users need to choose whether to use the input graph as directed or undirected. 
In Figure \ref{fg:01_year}, since the prediction target is the publication years of papers, edge directions play a crucial role, i.e., edge directions indicate time directions cited from new papers to old ones. 
For instance, nodes $6$ and $11$ have no incoming edge, and thus methods using directed graphs can predict that they are recent papers, i.e., their labels are $2022$. 
If we choose the input graph as undirected, methods cannot utilize the time direction. 
We also give another example in which an undirected graph is preferable in Figure \ref{fg:01_field}.
In this example, the nodes in a densely connected subgraph have the same label. 
For instance, node $17$ does not receive any information from its neighbor since it has no incoming edge.  
This means that methods using only directed graphs fail to utilize the existence of edges. 
Hence, users need to adaptively select a directed or undirected graph depending on the characteristics of datasets.

In summary, each combination of node representations and edge direction awareness exhibits the distinct capability, motivating us to develop a new method that adaptively captures all combinations. 

\section{Related Work}
\label{sec:related}
To address the node classification task, existing studies have proposed a variety of methods including GNNs and methods using adjacency lists. 
We briefly explain their motivations and several key instances. 

\smallskip \noindent \textbf{GNNs using feature aggregation in undirected graphs. }
Feature aggregation is an effective approach to obtain node representations denoised by locally neighboring nodes, e.g., the average, sum, or max pooling of neighboring nodes, specifically for homophilous graphs.
Feature aggregation is operated in graph convolution which has been proposed in \cite{kipf2017semi}. 
Since then, many successor GNN techniques \cite{monti2017geometric,hamilton2017inductive,velickovic2018graph,klicpera2018predict,xu2018powerful,zhu2020beyond,sign_icml_grl2020,zhu2021graph,shadow_GNN,maurya2021improving,chien2021adaptive} adopt feature aggregation in order to obtain node representations/embeddings suitable for various downstream tasks including node classification and link prediction. 
For example, SGC \cite{wu2019simplifying} simplifies GCN by removing nonlinearities between convolutional layers. SGC exhibits comparable performance to GCN while being computationally more efficient and fitting fewer parameters. 
Removing the nonlinearities between layers allows for feature aggregation to be performed prior to training learnable weights, enhancing the model's efficiency. This type of model is referred to as precomputation-based GNNs.
GPRGNN \cite{chien2021adaptive} automatically learns generalized PageRank parameters associated with each step of feature propagation. The parameters depend on the contributions of different steps during the information propagation. 
FSGNN \cite{maurya2021improving} is a precomputation-based GNN that flexibly utilizes information from high-order neighbors while reducing the number of model parameters by decoupling the depth of feature propagation and the number of layers of neural networks. 
ACMGCN \cite{luan2022revisiting} applies both low-pass and high-pass filters for each node in a layer, and adaptively fuses the generated node embeddings from each filter.  
These GNNs do not directly use adjacency lists as node features and focus on only undirected graphs. 

\smallskip \noindent \textbf{GNNs for directed graphs. }
While many studies have addressed developing techniques for undirected graphs, 
several studies \cite{tong2020directed,tong2020digraph,zhang2021magnet,ma2019spectral} have proposed techniques for node classification in directed graphs. 
DiGraph \cite{tong2020digraph} extends the spectral-based graph convolution to directed graphs by leveraging the inherent connections between graph Laplacian and stationary distributions of PageRank \cite{page1998pagerank}. 
MagNet \cite{zhang2021magnet} utilizes a complex Hermitian matrix known as the magnetic Laplacian, where the real and imaginary parts represent the undirected and directed edges, respectively.
However, similar to GNNs for undirected graphs, these GNNs for directed graphs do not use adjacency lists as node features.
While not only for node classification, several studies \cite{he2022gnnrank,Virinchi2023} have addressed a recommendation task in directed graphs, they are out of our scope. 

\smallskip \noindent \textbf{Methods using adjacency lists as node features. }
Several node embedding methods~\cite{huang2017accelerated,ye2018deep,CDE2018} leverage an adjacency matrix as node features. 
LINKX~\cite{lim2021large} achieves state-of-the-art performance on non-homophilous graphs. 
However, LINKX obtains lower accuracy on homophilous graphs than existing GNNs since it does not adopt graph convolution that is tailored to capture the homophily. 
A recent work, GloGNN++~\cite{li2022finding}, creates node features by combining an adjacency matrix and an attribute matrix. Then, it aggregates information from global nodes in the graph, i.e., nodes having similar combined node features, while other GNNs aggregate information from local neighbors\footnote{As we explained in this subsection, we categorize GloGNN++ into methods using adjacency lists since it utilizes adjacency lists for node features and does not use local feature aggregation.}. 
Since these methods use adjacency lists as node features, they can be easily applied to directed graphs by replacing undirected adjacency lists with directed adjacency lists. 
However, users need to decide whether to use a directed or undirected graph for each dataset. 
Also as for directed graphs, inverse edges may have different semantics from the original edges. 
Hence, these choices for an appropriate input to their methods increase users' burdens.
In summary, no existing methods combine aggregated features and adjacency lists in directed/undirected graphs though they complement each other.

\smallskip\noindent\textbf{Empirical studies on GNNs. }
Several studies \cite{NEURIPS2022_23ee05bf,dwivedi2020benchmarking,you2020design} empirically evaluate various GNNs to benchmark their performance from diverse perspectives such as scalability to large-scale graphs, accuracy in various tasks, and the impact of components in GNN architectures.
Other studies \cite{palowitch2022graphworld,maekawa2022beyond} aim to clarify the strengths and weaknesses of GNNs by generating various synthetic graphs.
To the best of our knowledge, no study has comprehensively explored both GNNs and methods utilizing adjacency lists with respect to edge direction awareness.

\section{A2DUG}
\label{sec:proposal}

This section presents a simple yet holistic method leveraging \textbf{A}ggregated features and \textbf{A}djacency lists in \textbf{D}irected/\textbf{U}ndirected \textbf{G}raphs, called \textbf{\textsc{A2DUG}}. 
In the following, we explain our design policy, method, concrete algorithm, relationships between our and existing methods, and complexity.

\subsection{Design Policy}
As discussed in Section \ref{ssec:motivating}, each combination of node representations (aggregated features vs. adjacency lists) and edge direction awareness (directed vs. undirected) holds unique distinguishability distinctions for the node classification task. 
This motivates us to adaptively control their effects as needed for each graph. 
Also, in practical contexts, the scalability of methods emerges as a critical concern, potentially affecting their suitability for real-world applications.
Thus, our focus is on addressing both effectiveness and scalability.
The model design is kept as simple as possible since this paper is the first study to leverage all combinations of node representations in directed and undirected graphs.
Therefore, A2DUG consists of simple neural network models instead of complex models, while it is modular; we can replace simple models to other GNN models.

\subsection{Method}

We describe \textbf{\textsc{A2DUG}} that satisfies two key design criteria for the effectiveness and scalability: 
\textbf{D1} leveraging both aggregated and adjacency lists features in directed/undirected graphs, \textbf{D2} adopting precomputation-based GNNs for minibatch training.

\smallskip \noindent \textbf{D1: Leveraging both aggregated features and adjacency lists in directed/undirected graphs.}
First, to adaptively control the effect from aggregated features and adjacency lists in directed/undirected graphs during the model training, we simply input all the combinations to multi-layer perceptions (MLPs) as follows:
\begin{align}
\label{eq:prediction}
    \bmY = \text{MLP}_\text{final}\Bigl(\sigma(\bmH_{\bmX}\|\bmH_{\bmA}\|\bmH_{\bmA^{\top}}\|\bmH_{\bmA^\text{und}}\nonumber\\ \|\bmH_\text{GNN}\|\bmH_{\text{GNN}^{\top}}\|\bmH_{\text{GNN}^\text{und}})\Bigr),
\end{align}
where $\bmY$ is the predicted labels, $\|$ indicates a concatenation operation, $\sigma$ indicates an activation function, e.g., ReLU, and $\bmH_{\bmX},\bmH_{\bmA},\allowbreak\bmH_{\bmA^{\top}},\allowbreak\bmH_{\bmA^\text{und}},\bmH_\text{GNN},\bmH_{\text{GNN}^{\top}},\bmH_{\text{GNN}^\text{und}}$ are node representation matrices of node features, adjacency lists, transposed adjacency lists, undirected adjacency lists, aggregated features in directed graphs, transposed aggregated features in directed graphs, and aggregated features in undirected graphs, respectively.
We include the node representations from \textit{transposed} directed graphs\footnote{Transposed directed graphs indicate graphs where edge directions are opposite to their original graphs.}, i.e., $\bmH_{\bmA^{\top}}$ and $\bmH_{\text{GNN}^{\top}}$, since inverse edges may have different semantics from the original edges, e.g., in citation networks, the reverse edge traces the descendant papers while the original edge traces the ancestor papers (see Section \ref{ssec:ablation} for more detailed discussion).
Each $\bmH$ is formulated as follows:
\begin{align}
\label{eq:h_components}
    &\bmH_{\bmX} = \text{MLP}_{\bmX}(\bmX),
    \bmH_{\bmA}\! =\! \text{MLP}_{\bmA}(\bmA),\nonumber \\ 
    &\bmH_{\bmA^{\top}}\! =\! \text{MLP}_{\bmA^{\top}}(\bmA^{\top}),
    \bmH_{\bmA^\text{und}}\! =\! \text{MLP}_{\bmA^\text{und}}(\bmA^\text{und}), \nonumber\\ 
    &\bmH_\text{GNN} = \text{GNN}(\bmA,\bmX), \ \ 
    \bmH_{\text{GNN}^{\top}} = {\text{GNN}^{\top}}(\bmA^{\top},\bmX), \nonumber \\ 
    &\bmH_{\text{GNN}^\text{und}} = \text{GNN}^\text{und}(\bmA^\text{und},\bmX), 
\end{align}
where $\text{GNN}, \text{GNN}^{\top},$ and $\text{GNN}^\text{und}$ are GNN-based encoders.
Since A2DUG leverages both aggregated features and adjacency lists in directed/undirected graphs,
it satisfies the first design criteria. 
While other model architectures may satisfy our design criteria, our model architecture is the simplest among the ones satisfying the design criteria. 
To show the validity of our simple model design, we compare A2DUG with its extension that has learnable layer selection parameters controlling the effects from node representations in Section \ref{ssec:model_validity}.

\smallskip \noindent \textbf{D2: Adopting precomputation-based GNNs for minibatch training.}
Second, to ensure the scalability of our proposal, we adopt precomputation-based GNNs, which ensures high scalability while showing competitive classification quality to other GNNs\footnote{Though other GNNs can also be applied to our proposal, its scalability cannot be ensured in the case.}. 
Existing precomputation-based GNNs \cite{wu2019simplifying,sign_icml_grl2020,maurya2021improving} compute feature aggregation as a preprocess. 
Their models thus can be trained with small input batches of node features once features are computed, leading to their high scalability.
Also, motivated by existing GNNs \cite{maurya2021improving}, we utilize both adjacency matrices with and without self-loops, i.e., $\bmA$ and $\hat{\bmA} = \bmA + \bmI$, which improve the classification quality in both homophilous and non-homophilous graphs. 
We formulate a $k$ layer GNN-based graph encoder as follows:
\begin{align}
\label{eq:gnn_dir}
    \text{GNN}_\text{precomp} = \text{MLP}(\bmA\bmX\|\hat{\bmA}\bmX\|\bmA^2\bmX\|\hat{\bmA}^2\bmX\nonumber\\ 
    \|\dots\|\bmA^k\bmX\|\hat{\bmA}^k\bmX). 
\end{align}
Also, we derive the formula for $\text{GNN}^{\top}_\text{precomp}$ by replacing $\bmA$ in Eq~\eqref{eq:gnn_dir} with $\bmA^{\top}$. 
It is worth noting that A2DUG can utilize high-order structural information similarly to other GNNs because it incorporates aggregated features through $k$ layers into its formulation. 

As for feature aggregation in an undirected graph, an adjacency matrix is typically normalized by node degrees \cite{kipf2017semi,velickovic2018graph,wu2019simplifying,maurya2021improving}. 
Following this, we apply the node degree-normalization to an undirected adjacency matrix and formulate a GNN-based graph encoder for undirected graphs as follows:
\begin{align}
\label{eq:gnn_und}
    \text{GNN}^\text{und}_\text{precomp} = \text{MLP}(\bmS\bmX\|\hat{\bmS}\bmX\|\bmS^2\bmX\|\hat{\bmS}^2\bmX\nonumber\\ 
    \|\dots\|\bmS^k\bmX\|\hat{\bmS}^k\bmX),
\end{align}
where $\bmS$ is a degree-normalized adjacency matrix, i.e., \\ $\bmS = (\bmD^\text{und})^{-\frac{1}{2}}\bmA^\text{und}(\bmD^\text{und})^{-\frac{1}{2}}$, and $\hat{\bmS}$ is a degree-normalized adjacency matrix with self-loops. 

\subsection{Training procedure}
We train our proposed model in an end-to-end training framework for a node classification task. 
To scale the training of the model for large-scale graphs, we adopt batchwise training which is introduced in existing precomputation-based GNNs \cite{wu2019simplifying,maurya2021improving} and methods using adjacency lists \cite{lim2021large}. 
To be concrete, we can handle $\bmX, \bmA, \bmA^{\top}, \bmA^\text{und}, \bmA\bmX, \hat{\bmA}\bmX,\bmA^{\top}\bmX,\allowbreak \hat{\bmA}^{\top}\bmX,\allowbreak\bmS\bmX,\allowbreak \hat{\bmS}\bmX,\allowbreak\dots,\allowbreak \bmA^k\bmX,\allowbreak  \hat{\bmA}^k\bmX,(\bmA^{\top})^k\bmX,\allowbreak  (\hat{\bmA}^{\top})^k\bmX, \allowbreak \bmS^k\bmX,$ and $\hat{\bmS}^k\bmX$ as node features and execute horizontal decomposition for batchwise training. 
Hence, the proposed method satisfies the second key design criteria, i.e., it scales well to large-scale graphs. 

\subsection{Algorithm}
We show the algorithm of A2DUG based on Eq. \eqref{eq:prediction}, \eqref{eq:h_components}, \eqref{eq:gnn_dir}, and \eqref{eq:gnn_und} in Algorithm \ref{al:A2DUG}, where $T$ is the number of training epochs. 
First, an undirected adjacency matrix is obtained from a given adjacency matrix in line 1. 
In lines 2--3, A2DUG computes adjacency matrices with self-loops and degree-normalized adjacency matrices.  
Then, in lines 5--11, node representations are obtained using MLPs that take as input node features, adjacency lists, and aggregated features. 
Notably, since all the computations of the aggregated features are independent of weight matrices in MLPs, the computations are performed only once as a preprocessing step before training iterations. 
In line 12, $\text{MLP}_\text{final}$ combines the node representations and then outputs the predicted labels. 
In our practical implementation, we employ horizontal decomposition to split the matrices into train/val/test data based on a given data split before training iterations. 

\begin{algorithm}[t]
\small
    \caption{A2DUG}
    \label{al:A2DUG}
        \DontPrintSemicolon
            \SetKwInOut{Input}{input}
            \SetKwInOut{Output}{output}
            \SetKwFunction{Expand}{Expand}
            \Input{$\bmA, \bmX, \bmC, k, T$} 
            \Output{$\bmY$}
            $\bmA^\text{und} = \text{to\_undirected}(\bmA)$ \hfill $\vartriangleright$ Obtaining undirected graph\\
            $\hat{\bmA} = \bmA + \bmI$, $\hat{\bmA}^\text{und} = \bmA^\text{und} + \bmI$ \hfill$\vartriangleright$ Adding self-loops\\
            $\bmS\! =\! (\bmD^\text{und})^{-\frac{1}{2}}\bmA^\text{und}(\bmD^\text{und})^{-\frac{1}{2}}, \hat{\bmS}\! =\! (\hat{\bmD}^\text{und})^{-\frac{1}{2}}\hat{\bmA}^\text{und}(\hat{\bmD}^\text{und})^{-\frac{1}{2}}$ \\ 
            \For{$i=1$ to $T$}{
                $\bmH_{\bmX} = \text{MLP}_{\bmX}(\bmX)$\\
                $\bmH_{\bmA} = \text{MLP}_{\bmA}(\bmA)$\\
                $\bmH_{\bmA^{\top}} = \text{MLP}_{\bmA^{\top}}(\bmA^{\top})$ \\
                $\bmH_{\bmA^\text{und}} = \text{MLP}_{\bmA^\text{und}}(\bmA^\text{und})$\\
                $\bmH_\text{GNN} = \text{MLP}_\text{GNN}(\bmA\bmX\|\hat{\bmA}\bmX\|\dots\|\bmA^k\bmX\|\hat{\bmA}^k\bmX)$\\
                $\bmH_{\text{GNN}^{\top}}= \text{MLP}_{\text{GNN}^{\top}}(\bmA^{\top}\bmX\|\hat{\bmA}^{\top}\bmX\|\dots\|(\bmA^{\top})^k\bmX\|(\hat{\bmA}^{\top})^k\bmX)$\\
                $\bmH_{\text{GNN}^\text{und}}= \text{MLP}_{\text{GNN}^\text{und}}(\bmS\bmX\|\hat{\bmS}\bmX\|\dots\|\bmS^k\bmX\|\hat{\bmS}^k\bmX)$\\
                $\bmY = \text{MLP}_\text{final}\Bigl(\sigma(\bmH_{\bmX}\|\bmH_{\bmA}\|\bmH_{\bmA^{\top}}\|\bmH_{\bmA^\text{und}}$ $\|\bmH_\text{GNN}\|\bmH_{\text{GNN}^{\top}}\|\bmH_{\text{GNN}^\text{und}})\Bigr)$\\
                Update weight matrices to optimize $\text{CrossEntropy}(\bmC, \bmY)$\\
                Check early stop criteria\\
            }
\end{algorithm} 

\begin{table*}[t]
    \caption{Summary of datasets.
    Datasets such as genius and pokec have relatively high $H(G)$ due to a small number of classes and/or imbalance classes while they do not exhibit the strong homophily property \cite{lim2021large}.
    We provide a code calculating $H(G)$ in our codebase for the reproducibility. 
    }
    \centering\scalebox{1}{
    \begin{tabular}{l|rrrrrrrc} \toprule
        Dataset & Nodes & Edges & Undirected Edges & Attributes & Labels & Prediction Target & $H(G)$ & Homophilous? \\\midrule \midrule
        cornell & $183$ & $298$ & $280$ & $1,703$ & $5$ & web page category & $0.131$ & \\
        texas & $183$ & $325$ & $295$ & $1,703$ & $5$ & web page category & $0.108$ &\\
        wisconsin & $251$ & $515$ & $466$ & $1,703$ & $5$ & web page category & $0.196$ & \\
        citeseer & $3,312$ & $4,715$ & $4,660$ & $3,703$ & $6$ & research field & $0.746$ & \checkmark\\
        coraML & $2,995$ & $8,416$ & $8,158$ & $2,879$ & $7$ & research field  & $0.792$ & \checkmark\\
        chameleon & $890$ & $13,584$ & $8,904$ & $2,325$ & $5$ & web page traffic & $0.247$ & \\
        squirrel & $2,223$ & $65,718$ & $47,138$ & $2,089$ & $5$ & web page traffic & $0.217$ & \\ \midrule
        genius & $421,961$ & $984,979$ & $922,868$ & $12$ & $2$ & marked act. & $0.618$ &\\
        arxiv-year & $169,343$ & $1,166,243$ & $1,157,799$ & $128$ & $5$ & publication year& $0.222$ &\\
        ogbn-arxiv& $169,343$ & $1,166,243$ & $1,157,799$ & $128$ & $40$ & research field& $0.655$ &  \checkmark\\ \midrule
        snap-patents & $2,923,922$ & $13,975,788$ & $13,972,547$ & $269$ & $5$ & time granted & $0.219$ &  \\
        pokec & $1,632,803$ & $30,622,564$ & $22,301,964$ & $65$ & $2$ & gender& $0.445$\\
        wiki & $1,925,342$ & $303,434,860$ & $242,605,360$ & $600$ & $5$ & total page views & $0.389$ &\\
        \bottomrule
    \end{tabular}
    }
\label{tb:04_dataset}
\end{table*}

\subsection{Connection with Existing Methods}
Since A2DUG is based on a general and holistic form of leveraging both aggregated features and adjacency lists in directed/undirected graphs, 
it can employ any existing method as a component in Eq. \eqref{eq:prediction}.
For example, A2DUG can leverage LINKX and GloGNN++, which are methods using adjacency lists, as one of the node representations in Eq. \eqref{eq:prediction}, e.g., $\bmH_{\bmA}$ or $\bmH_{\bmA^\text{und}}$. 
Other examples are GCN \cite{kipf2017semi}, FSGNN \cite{maurya2021improving}, and ACMGCN \cite{luan2022revisiting}, which can be used as $\bmH_{\text{GNN}^\text{und}}$. 
In this paper, to ensure high scalability, A2DUG adopts simple and scalable components in its model architecture (see Eq. \eqref{eq:h_components}, \eqref{eq:gnn_dir}, and \eqref{eq:gnn_und}). 

\subsection{Complexity}
The precomputation and training steps of A2DUG have time complexity of $\mathcal{O}(dk|E|+(h|E|+ndkh+nh^2L)T)$, in which $h$ is the hidden dimension, $|E|$ is the number of edges, $L$ is the number of MLP's layers, and $T$ is the number of epochs.
This is because it requires $\mathcal{O}(dk|E|)$ to precompute aggregated features within $k$-hops. 
Then, for the training step, it requires $\mathcal{O}(h|E|)$ cost for the first linear mapping of $\bmA, \bmA^{\top}$, and $\bmA^\text{und}$, $\mathcal{O}(ndkh)$ cost for the first linear mapping of aggregated features within $k$-hops, and $\mathcal{O}(nh^2L)$ cost for MLP operations on hidden features.
This complexity is comparable with existing scalable methods using aggregated features or adjacency lists, e.g., FSGNN or LINKX, since A2DUG also employs their efficient feature precomputation and model architectures. 

We provide an in-depth comparison of A2DUG against other scalable methods utilizing aggregated features or adjacency lists, e.g., FSGNN or LINKX.
First, FSGNN has the time complexity of $\mathcal{O}(dk|E|+(ndkh+nkh^2+nh^2L)T)$. 
The precomputation of feature aggregation requires $\mathcal{O}(dk|E|)$ cost and the training step requires $\mathcal{O}(ndkh)$ cost for the first linear mapping of aggregated features, $\mathcal{O}(nkh^2)$ cost for the mapping of the concatenated features, and $\mathcal{O}(nh^2L)$ for MLP operations on hidden features. 
The major difference from that of A2DUG is that it additionally requires $\mathcal{O}(h|E|T)$ cost for the linear mapping of adjacency lists, which is not a significant cost. 

Next, LINKX has the time complexity of $\mathcal{O}((h|E|+ndh+nh^2L)T)$. 
It requires $\mathcal{O}(d|E|)$ cost for the first linear mapping of adjacency lists, $\mathcal{O}(ndh)$ cost for the first linear mapping of node features, and $\mathcal{O}(nh^2L)$ cost for MLP operations on hidden features. 
Hence, LINKX has a similar time complexity to the training of A2DUG since they share the MLP-based model architecture.
Compared with LINKX, A2DUG additionally require $\mathcal{O}(dk|E|)$ for the precomputation of aggregated features. 
In practice, the precomputation is efficient because it is executed only once before training. 

Finally, we discuss the computational benefits of the design choices of A2DUG. 
Adding more layers to MLPs only gives $\mathcal{O}(h^2)$ and thus does not scale to the number of edges. 
This is because the graph information in $\bmA, \bmA^{\top}$, and $\bmA^\text{und}$ is embedded into hidden vectors after the first linear mapping of MLPs. 
Further, this enables a sparse-dense matrix product to compute the first linear mappings of $\text{MLP}_{\bmA}, \text{MLP}_{\bmA^{\top}}$, and $\text{MLP}_{\bmA^\text{und}}$, which largely improves the efficiency since real-world graphs are typically very sparse. 
\section{Experiments}
\label{sec:experiments}

We aim to answer the following questions: 
\begin{itemize}
    \item[\textbf{Q1.}] \textit{How effectively does A2DUG perform on various directed graphs?} 
    \item[\textbf{Q2.}] \textit{How does A2DUG scale well?} 
    \item[\textbf{Q3.}] \textit{To what extent do aggregated features and adjacency lists in directed/undirected graphs affect node classification?}
    \item[\textbf{Q4.}] \textit{Is the model design of A2DUG valid?}
\end{itemize}

\smallskip \noindent \textbf{Datasets.}
To comprehensively evaluate A2DUG with existing methods, we test them on 13 directed graphs with varying the graph sizes.
Table \ref{tb:04_dataset} shows the statistics of datasets.
We use seven small-scale directed graphs:
\begin{itemize}
    \item \texttt{cornell}, \texttt{texas}, and \texttt{wisconsin}\footnote{\url{http://www.cs.cmu.edu/afs/cs.cmu.edu/project/theo-11/www/wwkb/}} are graphs representing links between
web pages of the corresponding universities. The task is to classify the nodes into one of the five categories, student, project, course, staff, and faculty.
    \item \texttt{citeseer} and \texttt{coraML}\footnote{We use the versions
of these datasets provided in \cite{zhang2021magnet}.} are widely used citation networks. The task is to predict the research fields of nodes. 
    \item \texttt{chameleon} and \texttt{squirrel}~\cite{platonov2023a} are Wikipedia networks on specific topics. The task is to predict the page views. Since a study \cite{platonov2023a} pointed out the presence of a large number of duplicate nodes in the original datasets of \texttt{chameleon} and \texttt{squirrel}, we use the filtered versions that do not have any duplicate nodes.
\end{itemize}
We use three middle-scale directed graphs:
\begin{itemize}
    \item \texttt{genius} \cite{lim2021large} is a subset of the social network on genius.com. Nodes are users, and edges connect users that follow each other on the site. The task is to predict whether nodes are marked ``gone'', which appears to often include spam users.
    \item \texttt{arxiv-year} \cite{lim2021large} is a citation network, where the nodes are arXiv papers and directed edges connect a paper to other papers that it cites. The task is to predict the year at which the paper is posted.
    \item \texttt{ogbn-arxiv} \cite{hu2020ogb} has the same graph structure and node features to \texttt{arxiv-year} but has different classes that indicate the research fields of papers. 
\end{itemize}
Also, we use three large-scale directed graphs: 
\begin{itemize}
    \item \texttt{pokec} \cite{lim2021large} is the friendship graph of a Slovak online social network, where nodes are users and edges are directed friendship relations. Nodes are labeled with reported gender.
    \item \texttt{snap-patents} \cite{lim2021large} is a dataset of utility patents in the US. Each node is a patent, and edges connect patents that cite each other. The task is to predict the time at which a patent was granted. 
    \item \texttt{wiki} \cite{lim2021large} is a dataset of Wikipedia articles, where nodes represent pages and edges represent links between them. The task is to predict page views. 
\end{itemize}

\smallskip\noindent\textbf{Data split. }
We follow the same way to split training/validation/test sets as the papers that proposed the datasets \cite{hu2020ogb,lim2021large,platonov2023a}.
To be concrete, for cornell, texas, and wisconsin, we use the default split (48/32/20 train/val/test) provided in \textit{PyG}\footnote{\url{https://github.com/pyg-team/pytorch_geometric}}. 
For chameleon and squirrel, the study \cite{platonov2023a} that has proposed them provides $10$ different splits. Hence, we utilize five different splits from them in our experiments. 
As for citeseer and coraML, following the rules in \cite{zhang2021magnet}, we choose 20 labels per class for the training set, 500 labels for the validation set, and the rest for the test set. 
As for ogbn-arxiv, the study \cite{hu2020ogb} that has proposed them provides a single data split. 
Hence, we simply utilize it and initialize model parameters by using different random seeds for each run. 
For genius, snap-pantents, pokec, and wiki, the study \cite{lim2021large} that has proposed them  run each method on the same five random 50/25/25 train/val/test splits for each dataset. 
We follow the way to split the datasets. 

\begin{table*}[t]
\caption{Experimental results. Test accuracy is displayed for most datasets, while genius displays test ROC AUC. Standard deviations are calculated from five runs using different random seeds. The two best results per dataset are highlighted. 
In the bottom row, we show the accuracy difference between A2DUG and the best existing method for each dataset. 
(M) denotes some (or all) hyperparameter settings run out of memory and (TO) denotes that the runs do not finish in 24 hours. We show the average rank of each method across all datasets. 
}
\centering
\scalebox{.96}{
\begin{tabular}{l|ccccccc}
\toprule
 & cornell & texas & wisconsin & citeseer & coraML & chameleon & squirrel \\
\midrule
MLP & $75.68_{\pm0.00}$ & $73.51_{\pm3.52}$ & $75.29_{\pm1.75}$ & $52.89_{\pm2.76}$ & $62.23_{\pm2.76}$ & $37.65_{\pm3.10}$ & $35.13_{\pm3.87}$ \\\midrule
GCN & $42.16_{\pm3.08}$ & $51.89_{\pm3.52}$ & $51.76_{\pm2.24}$ & \colorbox{mycolor}{$65.88_{\pm1.50}$} & \colorbox{mycolor}{$79.27_{\pm2.85}$} & $37.17_{\pm3.37}$ & $32.26_{\pm2.96}$ \\
SGC & $41.62_{\pm1.48}$ & $59.46_{\pm0.00}$ & $55.29_{\pm0.88}$ & $62.61_{\pm2.19}$ & $78.68_{\pm2.87}$ & $38.64_{\pm3.79}$ & $38.50_{\pm1.93}$ \\
GPRGNN & $63.24_{\pm3.08}$ & $82.16_{\pm4.10}$ & $74.90_{\pm2.91}$ & $63.00_{\pm1.55}$ & $77.96_{\pm2.75}$ & $38.22_{\pm5.05}$ & $35.03_{\pm1.46}$ \\
FSGNN & \colorbox{mycolor}{$76.22_{\pm1.21}$} & \colorbox{mycolor}{$83.24_{\pm2.26}$} & $76.47_{\pm1.39}$ & \colorbox{mycolor}{$68.56_{\pm0.72}$} & \colorbox{mycolor}{$81.11_{\pm2.05}$} & \colorbox{mycolor}{$44.96_{\pm1.02}$} & \colorbox{mycolor}{$41.12_{\pm2.58}$} \\
ACMGCN & $60.54_{\pm8.02}$ & $71.89_{\pm4.10}$ & $65.88_{\pm5.48}$ & $64.18_{\pm2.31}$ & $77.03_{\pm3.71}$ & $39.44_{\pm4.92}$ & $38.85_{\pm0.56}$ \\\midrule
LINK undirected & $56.76_{\pm4.27}$ & $73.51_{\pm2.26}$ & $41.57_{\pm2.56}$ & $28.67_{\pm4.78}$ & $51.24_{\pm3.37}$ & $41.60_{\pm5.62}$ & $36.31_{\pm1.90}$ \\
LINKX undirected & \colorbox{mycolor}{$76.76_{\pm3.08}$} & $75.14_{\pm4.01}$ & $75.69_{\pm1.75}$ & $47.18_{\pm5.13}$ & $70.10_{\pm1.69}$ & $41.55_{\pm2.45}$ & $40.10_{\pm2.64}$ \\
GloGNN++ undirected & $75.14_{\pm2.26}$ & $78.38_{\pm0.00}$ & \colorbox{mycolor}{$79.22_{\pm1.07}$} & $53.50_{\pm2.55}$ & $63.43_{\pm6.56}$ & $38.49_{\pm9.02}$ & $34.48_{\pm7.13}$ \\\midrule
LINK directed & $49.19_{\pm2.26}$ & $71.89_{\pm3.63}$ & $50.20_{\pm5.48}$ & $20.58_{\pm1.86}$ & $51.67_{\pm5.59}$ & $38.21_{\pm2.82}$ & $33.06_{\pm2.62}$ \\
LINKX directed & $75.14_{\pm2.26}$ & $73.51_{\pm2.26}$ & $77.65_{\pm2.24}$ & $52.51_{\pm1.40}$ & $59.42_{\pm7.85}$ & $39.64_{\pm5.95}$ & $36.00_{\pm1.18}$ \\
GloGNN++ directed & $64.86_{\pm14.93}$ & $78.38_{\pm0.00}$ & \colorbox{mycolor}{$80.39_{\pm0.00}$} & $46.31_{\pm15.95}$ & $69.02_{\pm1.48}$ & $40.07_{\pm3.36}$ & $34.87_{\pm8.12}$ \\ \midrule
DGCN & $56.76_{\pm8.96}$ & $67.03_{\pm7.74}$ & $54.90_{\pm2.40}$ & $64.65_{\pm3.41}$ & $78.05_{\pm3.00}$ & $42.42_{\pm4.60}$ & $41.04_{\pm2.07}$ \\
Digraph & $47.03_{\pm3.63}$ & $42.70_{\pm4.83}$ & $50.98_{\pm0.00}$ & $61.41_{\pm1.95}$ & $73.62_{\pm4.78}$ & $34.29_{\pm0.82}$ & $35.70_{\pm1.60}$ \\
DigraphIB & $48.65_{\pm2.70}$ & $61.08_{\pm4.52}$ & $58.04_{\pm2.63}$ & $59.40_{\pm1.80}$ & $74.62_{\pm3.63}$ & $38.28_{\pm2.69}$ & $33.90_{\pm2.04}$ \\
Magnet & $71.89_{\pm3.08}$ & \colorbox{mycolor}{$83.24_{\pm7.97}$} & $72.94_{\pm3.22}$ & $55.39_{\pm9.23}$ & $76.93_{\pm4.44}$ & $35.16_{\pm3.38}$ & $31.37_{\pm1.46}$ \\ \midrule
\textbf{A2DUG} & $74.59_{\pm1.48}$ & \colorbox{mycolor}{$84.32_{\pm2.96}$} & $77.65_{\pm1.75}$ & $64.55_{\pm3.13}$ & $77.64_{\pm1.71}$ & \colorbox{mycolor}{$42.78_{\pm4.79}$} & \colorbox{mycolor}{$42.28_{\pm2.36}$} \\
Gain over SOTA & $-2.17$ & $+1.08$ & $-2.74$ & $-4.01$ & $-3.47$ & $-2.18$ & $+1.16$ \\ \bottomrule
\end{tabular}
}
\vspace{3mm}

\scalebox{1.}{
\begin{tabular}{l|ccc|ccc|c}
\toprule

{}&genius&arxiv-year&ogbn-arxiv&snap-patents&pokec&wiki&Avg. rank\\
\midrule
MLP & $85.84_{\pm0.88}$ &$36.92_{\pm0.23}$ & $53.78_{\pm0.29}$ &$31.49_{\pm0.07}$ &$62.53_{\pm0.04}$ &$39.74_{\pm0.28}$ & $10.54$\\ \midrule
GCN & $82.23_{\pm3.42}$ &$43.73_{\pm0.22}$ & $69.30_{\pm0.16}$ &$39.99_{\pm0.35}$ & $63.05_{\pm4.23}$ & (M) & $10.38$\\
SGC & $80.08_{\pm2.82}$ &$38.79_{\pm0.22}$ & $67.40_{\pm0.02}$ &$35.26_{\pm0.04}$ &$69.83_{\pm0.32}$ &$45.07_{\pm0.09}$ & $9.77$\\
GPRGNN& $83.98_{\pm0.54}$ &$40.21_{\pm0.33}$ & $67.52_{\pm0.80}$ &$32.44_{\pm0.25}$ &$65.79_{\pm8.59}$ & (M) & $8.69$\\
FSGNN & $88.95_{\pm1.51}$ &$45.99_{\pm0.35}$ & \colorbox{mycolor}{$71.26_{\pm0.33}$} &$45.44_{\pm0.05}$ &$78.21_{\pm1.09}$ &$58.40_{\pm0.26}$ & $3.38$ \\
ACMGCN& $73.16_{\pm8.27}$ &$43.30_{\pm0.90}$ & $67.51_{\pm0.69}$ &$40.07_{\pm0.40}$ &$66.91_{\pm0.66}$ & (M) & $8.69$ \\\midrule
LINK undirected & $69.16_{\pm0.11}$ &$48.43_{\pm0.10}$ & $63.33_{\pm0.04}$ &$49.93_{\pm0.07}$ &$79.17_{\pm0.05}$ &$58.42_{\pm0.04}$ & $9.62$\\
LINKX undirected& $89.27_{\pm1.11}$ &$47.90_{\pm0.20}$ & $61.78_{\pm0.40}$ &$51.40_{\pm0.11}$ &$79.44_{\pm0.13}$ &$61.02_{\pm0.36}$ & $6.08$ \\
GloGNN++ undirected & \colorbox{mycolor}{$90.00_{\pm0.38}$} &$50.55_{\pm0.12}$ & $46.30_{\pm2.74}$ & (M) &\colorbox{mycolor}{$82.66_{\pm0.07}$} & (M) & $7.85$ \\ \midrule
LINK directed & $55.46_{\pm0.11}$ &$51.71_{\pm0.22}$ & $57.17_{\pm0.04}$ &$57.54_{\pm0.07}$ &$71.53_{\pm0.09}$ &$59.69_{\pm0.03}$ & $11.31$ \\
LINKX directed& $88.35_{\pm0.45}$ &$52.61_{\pm0.26}$ & $59.81_{\pm0.43}$ &\colorbox{mycolor}{$61.09_{\pm0.07}$} &$71.88_{\pm0.09}$ &\colorbox{mycolor}{$62.08_{\pm0.14}$} & $6.77$ \\
GloGNN++ directed & $87.82_{\pm0.13}$ &\colorbox{mycolor}{$53.67_{\pm0.32}$} &$55.36_{\pm20.59}$ & (M) &$75.36_{\pm0.07}$ & (M) & $7.54$ \\ \midrule
DGCN& (TO) & (TO) &(TO) &(M) & (M) &(M) & $10.15$\\
Digraph & (M) & (M) &(M) & (M) & (M) &(M) & $13.31$\\
DigraphIB & (M) & (M) &(M) & (M) & (M) &(M) & $12.62$\\
Magnet& $86.68_{\pm2.78}$ &$52.10_{\pm0.32}$ & $68.50_{\pm0.11}$ & (M) &$75.14_{\pm1.59}$ & (M) & $8.69$ \\ \midrule
\textbf{\textsc{A2DUG}}&\colorbox{mycolor}{$89.85_{\pm3.15}$}&\colorbox{mycolor}{$59.14_{\pm0.48}$}&\colorbox{mycolor}{$69.51_{\pm0.24}$}&\colorbox{mycolor}{$72.38_{\pm0.10}$}&\colorbox{mycolor}{$82.55_{\pm0.08}$}&\colorbox{mycolor}{$65.13_{\pm0.07}$} & $2.46$\\
Gain over SOTA & $-0.15$ & $+5.47$ & $-1.75$ & $+11.29$ & $-0.11$ & $+3.05$ &  \\ \bottomrule
\end{tabular}
}
\label{tb:accuracy}
\end{table*}

\smallskip \noindent \textbf{Baselines. }
For GNNs using feature aggregation in undirected graphs, we use GCN\footnote{\url{https://github.com/tkipf/pygcn}} \cite{kipf2017semi}, SGC\footnote{\url{https://github.com/Tiiiger/SGC}} \cite{wu2019simplifying}, 
GPRGNN\footnote{\url{https://github.com/jianhao2016/GPRGNN}} \cite{chien2021adaptive}, FSGNN\footnote{\url{https://github.com/sunilkmaurya/FSGNN}} \cite{maurya2021improving},
and ACMGCN\footnote{\url{https://github.com/SitaoLuan/ACM-GNN}} \cite{luan2022revisiting}. 
For GNNs for directed graphs, we use DGCN\cite{tong2020directed}, DiGraph \cite{tong2020digraph}, its variant DiGraphIB, and MagNet\footnote{\url{https://github.com/matthew-hirn/magnet}}~\cite{zhang2021magnet}. 
As for methods using adjacency lists as node features, we use LINK~\cite{zheleva2009join}, LINKX\footnote{\url{https://github.com/CUAI/Non-Homophily-Large-Scale}}~\cite{lim2021large}, and GloGNN++\footnote{\url{https://github.com/recklessronan/glognn}}~\cite{li2022finding}. 
We use their official implementations if they are publicly available\footnote{We used the code of LINK which was implemented in LINKX's repository and the codes of DGCN, DiGraph, and DiGraphIB which were implemented in Magnet's repository.}.
We also execute a graph-agnostic classifier, multi-layer perceptron (MLP) as a baseline, i.e., it ignores the topology structure. 
Though we drop several existing methods such as \cite{velickovic2018graph, xu2018powerful, hamilton2017inductive, sign_icml_grl2020, zhu2020beyond} due to the space limitation, we do not fail to use the state-of-the-art methods that show superior performance to them in the papers. 

\smallskip \noindent \textbf{Settings.} 
We report the performance as mean classification accuracy and standard deviation over five random runs with different random seeds. 
As following \cite{lim2021large}, we use ROC-AUC as the metric for the class-imbalanced genius dataset (about $80\%$ of nodes are in the majority class), as it is less sensitive to class-imbalance than accuracy. 
In the methods using adjacency lists (i.e., LINK, LINKX, and GloGNN++), we evaluate both directed and undirected graphs as their inputs\footnote{In their papers, the authors manually chose a directed or undirected graph for each dataset, which obtains better results than another. }.
Following \cite{lim2021large}, we use the AdamW optimizer for gradient-based optimization in all experiments of this paper. 

We use a single NVIDIA A100-PCIE-40GB for all our experiments.
In large-scale datasets, snap-patents, pokec, and wiki, we use minibatch training by setting batchsizes to $n/10, n/10$, and $n/20$, respectively, for SGC, FSGNN, LINK, LINKX, and A2DUG which support minibatch training. 
To efficiently precompute aggregated features on the largest dataset wiki, which does not fit into the GPU memory, we utilize the block-wise
precomputation scheme proposed in \cite{maekawa2022gnn}.

\smallskip\noindent\textbf{Hyperparameters. } 
According to the papers or codebases of existing methods, we select hyperparameters for the search space, e.g., weight decays, learning rates, the number of hidden units, and dropout ratio. 
For all experiments, we choose the best parameter set from these candidates by utilizing Optuna~\cite{akiba2019optuna} for $100$ trials. 
For the reproduction of all experiments in this section,
we report the best set of hyperparameters for each experiment in our codebase. 
Please see README in the codebase, to find the search space and best parameter sets.

\begin{table*}[t]
    \caption{Comparison on training time [s] (per epoch / total). Note that total training time includes precomputation time for SGC, FSGNN, and \textsc{A2DUG}. SGC, FSGNN, LINK, LINKX, and A2DUG use minibatch training on snap-patent, pokec, and wiki. (M) denotes some (or all) hyperparameter settings run out of memory and (TO) denotes that the runs do not finish in 24 hours. 
    }
    \centering
\scalebox{0.95}{
\begin{tabular}{l|ccccccc}
\toprule
 & cornell & texas & wisconsin & citeseer & coraML & chameleon & squirrel \\
\midrule
MLP & $0.01$ / $5.86$ & $0.01$ / $6.08$ & $0.01$ / $5.76$ & $0.02$ / $6.57$ & $0.02$ / $7.28$ & $0.02$ / $7.03$ & $0.03$ / $7.63$ \\ \midrule
GCN & $0.02$ / $5.92$ & $0.02$ / $5.92$ & $0.02$ / $6.08$ & $0.02$ / $6.20$ & $0.02$ / $6.33$ & $0.02$ / $7.58$ & $0.03$ / $9.25$ \\
SGC & $0.01$ / $5.72$ & $0.01$ / $5.92$ & $0.01$ / $6.02$ & $0.02$ / $6.60$ & $0.02$ / $7.06$ & $0.02$ / $7.34$ & $0.02$ / $8.44$ \\
GPRGNN & $0.02$ / $6.71$ & $0.02$ / $7.74$ & $0.02$ / $6.35$ & $0.03$ / $6.33$ & $0.03$ / $6.67$ & $0.03$ / $7.40$ & $0.03$ / $6.98$ \\
FSGNN & $0.03$ / $7.44$ & $0.03$ / $7.15$ & $0.03$ / $7.97$ & $0.05$ / $9.90$ & $0.04$ / $10.03$ & $0.05$ / $10.95$ & $0.06$ / $12.80$ \\
ACMGCN & $0.03$ / $7.46$ & $0.03$ / $6.73$ & $0.03$ / $7.18$ & $0.03$ / $9.33$ & $0.03$ / $8.35$ & $0.03$ / $8.74$ & $0.04$ / $9.70$ \\ \midrule
LINK undirected & $0.01$ / $6.14$ & $0.01$ / $6.45$ & $0.02$ / $5.88$ & $0.02$ / $6.32$ & $0.02$ / $6.30$ & $0.03$ / $8.09$ & $0.03$ / $8.65$ \\
LINKX undirected & $0.02$ / $6.50$ & $0.02$ / $6.41$ & $0.02$ / $6.40$ & $0.02$ / $6.76$ & $0.02$ / $6.59$ & $0.04$ / $9.09$ & $0.04$ / $10.07$ \\
GloGNN++ undirected & $0.02$ / $8.87$ & $0.02$ / $7.84$ & $0.03$ / $8.01$ & $0.03$ / $6.97$ & $0.04$ / $17.43$ & $0.03$ / $11.55$ & $0.04$ / $11.33$ \\ \midrule
LINK directed & $0.02$ / $6.12$ & $0.01$ / $6.39$ & $0.02$ / $6.07$ & $0.02$ / $6.13$ & $0.02$ / $6.61$ & $0.02$ / $6.70$ & $0.02$ / $7.39$ \\
LINKX directed & $0.02$ / $6.36$ & $0.02$ / $6.78$ & $0.02$ / $6.34$ & $0.02$ / $6.60$ & $0.02$ / $6.34$ & $0.03$ / $8.27$ & $0.03$ / $7.39$ \\
GloGNN++ directed & $0.03$ / $7.70$ & $0.02$ / $7.60$ & $0.03$ / $8.49$ & $0.03$ / $10.02$ & $0.03$ / $9.32$ & $0.03$ / $10.10$ & $0.04$ / $14.88$ \\ \midrule
DGCN & $0.03$ / $7.09$ & $0.03$ / $7.82$ & $0.03$ / $6.83$ & $0.03$ / $12.89$ & $0.03$ / $13.51$ & $0.03$ / $11.23$ & $0.04$ / $27.55$ \\
Digraph & $0.01$ / $6.26$ & $0.02$ / $5.89$ & $0.02$ / $6.60$ & $0.02$ / $6.33$ & $0.02$ / $6.45$ & $0.02$ / $6.56$ & $0.02$ / $7.59$ \\
DigraphIB & $0.02$ / $6.18$ & $0.02$ / $6.29$ & $0.02$ / $6.42$ & $0.02$ / $6.08$ & $0.02$ / $6.92$ & $0.02$ / $6.58$ & $0.03$ / $7.26$ \\
Magnet & $0.04$ / $8.20$ & $0.04$ / $7.34$ & $0.05$ / $8.20$ & $0.07$ / $9.19$ & $0.09$ / $10.99$ & $0.05$ / $8.72$ & $0.08$ / $9.51$ \\ \midrule
\textbf{A2DUG} & $0.04$ / $8.43$ & $0.04$ / $7.94$ & $0.04$ / $8.08$ & $0.08$ / $17.71$ & $0.07$ / $21.33$ & $0.08$ / $14.13$ & $0.13$ / $21.23$ \\
\bottomrule
\end{tabular}
}

\vspace{2mm}

\scalebox{0.95}{
\begin{tabular}{l|ccc|ccc}
\toprule

{}&genius&arxiv-year&ogbn-arxiv&snap-patents&pokec&wiki\\
\midrule
MLP & $1.14$ / $64.94$ & $0.28$ / $85.14$ &$0.43$ / $148.84$ &$4.79$ / $439.55$ & $5.37$ / $1862.87$ &$3.42$ / $711.87$ \\\midrule
GCN &$1.47$ / $162.67$ &$0.39$ / $217.17$ &$0.55$ / $264.27$ & $4.57$ / $2461.39$ & $4.42$ / $1894.87$ &(M) \\
SGC & $1.04$ / $87.46$ &$0.26$ / $101.32$ &$0.39$ / $218.24$ & $5.69$ / $1005.80$ & $4.80$ / $2593.40$ & $3.37$ / $1248.20$ \\
GPRGNN& $1.09$ / $70.06$ &$0.31$ / $107.89$ &$0.51$ / $122.57$ &$4.41$ / $628.06$ & $3.97$ / $1580.53$ &(M) \\
FSGNN &$1.28$ / $189.42$ &$0.44$ / $215.82$ &$0.58$ / $285.87$ & $8.88$ / $4868.19$ & $5.31$ / $2862.49$ &$10.65$ / $5555.48$ \\
ACMGCN& $1.34$ / $86.94$ &$0.46$ / $225.45$ &$0.63$ / $334.94$ & $4.35$ / $2369.63$ & $4.83$ / $2643.42$ &(M) \\\midrule
LINK undirected & $1.27$ / $91.71$ & $0.41$ / $25.79$ &$0.48$ / $32.94$ & $10.59$ / $508.43$ &$5.31$ / $251.76$ &$11.84$ / $1280.93$ \\
LINKX undirected&$1.53$ / $177.23$ & $0.50$ / $32.95$ &$0.56$ / $47.10$ & $12.92$ / $592.94$ &$5.65$ / $299.37$ &$13.29$ / $2005.72$ \\
GloGNN++ undirected &$1.46$ / $136.71$ & $0.48$ / $58.40$ & $0.81$ / $258.42$ &(M) &$5.47$ / $847.04$ &(M) \\ \midrule
LINK directed & $1.12$ / $68.16$ & $0.65$ / $38.64$ &$0.71$ / $50.14$ & $10.87$ / $719.02$ &$6.38$ / $300.13$ &$11.18$ / $1306.04$ \\
LINKX directed&$1.68$ / $209.04$ & $0.64$ / $63.96$ &$0.78$ / $51.99$ & $12.96$ / $656.74$ &$7.81$ / $421.38$ &$12.31$ / $1616.59$ \\
GloGNN++ directed &$1.70$ / $212.61$ & $0.33$ / $62.49$ &$0.63$ / $268.37$ & (M) & $5.04$ / $1221.02$ &(M) \\ \midrule
DGCN& (TO) & (TO) &(TO) &(M) & (M) &(M)\\
Digraph & (M) & (M) &(M) & (M) & (M) &(M)\\
DigraphIB & (M) & (M) &(M) & (M) & (M) &(M)\\
Magnet&$1.31$ / $159.55$ &$0.60$ / $233.67$ &$0.74$ / $876.22$ &(M) &$5.92$ / $10112.13$ &(M) \\\midrule
\textbf{A2DUG} &$1.43$ / $357.05$ &$0.78$ / $113.27$ &$0.81$ / $180.28$ &$27.97$ / $2997.52$ & $9.92$ / $3293.78$ &$50.53$ / $5129.92$ \\

\bottomrule
\end{tabular}
}
\label{tb:efficiency}
\end{table*}

\begin{table*}[t]
    \caption{Ablation study of A2DUG.
    We highlight the best score on each dataset. 
    }
    \centering

\begin{tabular}{l|ccccccc}
\toprule
 & cornell & texas & wisconsin & citeseer & coraML & chameleon & squirrel \\
\midrule
\textbf{A2DUG} & \colorbox{mycolor}{$74.59_{\pm1.48}$} & $84.32_{\pm2.96}$ & $77.65_{\pm1.75}$ & $64.55_{\pm3.13}$ & $77.64_{\pm1.71}$ & $42.78_{\pm4.79}$ & \colorbox{mycolor}{$42.28_{\pm2.36}$} \\\midrule
wo directed & $72.97_{\pm1.91}$ & $86.49_{\pm4.68}$ & $74.90_{\pm2.15}$ & \colorbox{mycolor}{$65.22_{\pm2.72}$} & \colorbox{mycolor}{$80.85_{\pm2.33}$} & $41.38_{\pm2.68}$ & $35.68_{\pm9.95}$ \\
wo undirected & $69.73_{\pm4.01}$ & \colorbox{mycolor}{$87.57_{\pm3.08}$} & \colorbox{mycolor}{$80.00_{\pm0.88}$} & $55.42_{\pm2.73}$ & $71.00_{\pm2.25}$ & $40.83_{\pm7.91}$ & $40.87_{\pm2.98}$ \\
wo aggregation & $72.43_{\pm7.74}$ & $77.84_{\pm2.96}$ & $76.47_{\pm3.67}$ & $54.70_{\pm1.93}$ & $71.85_{\pm1.45}$ & $43.45_{\pm1.68}$ & $40.82_{\pm1.80}$ \\
wo adjacency & $72.97_{\pm3.31}$ & $82.16_{\pm2.42}$ & $66.27_{\pm0.88}$ & $64.64_{\pm1.76}$ & $78.79_{\pm1.90}$ & \colorbox{mycolor}{$44.79_{\pm2.14}$} & $41.93_{\pm2.10}$ \\
wo transpose & $68.65_{\pm4.10}$ & $84.86_{\pm4.10}$ & $72.16_{\pm1.64}$ & $64.19_{\pm1.79}$ & $79.35_{\pm2.17}$ & $40.73_{\pm8.90}$ & $41.39_{\pm1.97}$ \\
\bottomrule
\end{tabular}

\vspace{3mm}

\begin{tabular}{l|ccc|ccc}
\toprule
{}&genius&arxiv-year&ogbn-arxiv&snap-patents&pokec&wiki\\
\midrule
\textbf{\textsc{A2DUG}}&$89.85_{\pm3.15}$&$59.14_{\pm0.48}$&$69.51_{\pm0.24}$&\colorbox{mycolor}{$ 72.38_{\pm0.10}$}&$82.55_{\pm0.08}$&\colorbox{mycolor}{$65.13_{\pm0.07}$}\\ \midrule
wo directed&\colorbox{mycolor}{$90.96_{\pm0.10}$}&$51.46_{\pm0.28}$&\colorbox{mycolor}{$70.54_{\pm0.12}$}&$55.09_{\pm0.10}$&\colorbox{mycolor}{$83.02_{\pm0.06}$}&$63.11_{\pm0.09}$\\
wo undirected&$89.32_{\pm1.17}$&\colorbox{mycolor}{$60.13_{\pm0.45}$}&$65.80_{\pm0.23}$&$71.60_{\pm0.09}$&$81.14_{\pm0.04}$&$64.33_{\pm0.19}$\\
wo aggregation&$89.24_{\pm0.32}$&$54.88_{\pm0.36}$&$66.64_{\pm0.05}$&$62.97_{\pm0.40}$&$81.18_{\pm0.07}$&$64.57_{\pm0.17}$\\
wo adjacency&$90.65_{\pm1.72}$&$56.88_{\pm0.19}$&$69.54_{\pm0.91}$&$70.24_{\pm0.16}$&$78.42_{\pm2.64}$&$57.42_{\pm1.82}$\\
wo transpose&$89.63_{\pm0.58}$&$57.31_{\pm0.29}$&$70.29_{\pm0.27}$&$69.09_{\pm0.15}$&$81.85_{\pm0.11}$&$64.81_{\pm0.19}$\\
\bottomrule
\end{tabular}
\label{tb:ablation}
\end{table*}

\subsection{Q1. Effectiveness}
\label{ssec:effectiveness}

\smallskip \noindent \textbf{A2DUG achieves promising results across all datasets. } 
To answer the first question, we benchmark the performance of existing methods and A2DUG in various directed graphs in Table \ref{tb:accuracy}. 
The table shows the performance stability of A2DUG in various datasets. 
Despite its simple architecture, \textsc{A2DUG} outperforms other state-of-the-art methods with large margins in arxiv-year, snap-patents, and wiki, while the relative performance drop over the best method is less than $6\%$. 
The obtained findings suggest the imperative need to control the effects stemming from aggregated features and adjacency lists in both directed/undirected graphs for optimal performance across the three datasets. 
Notably, A2DUG stands out as the sole method capable of adaptively controlling these effects, rather than merely selecting pivotal features for classification. 
Consequently, A2DUG demonstrates strong performance in this regard.

\noindent\textbf{No existing method stably achieves the state-of-the-art classification quality across various datasets. }
For example, FSGNN perform well on small-scale commonly used datasets but shows significant accuracy drops in arxiv-year, snap-patents, and wiki compared with the state-of-the-art performance. This indicates that aggregated features in undirected graphs are important signal for correctly predicting node labels in small-scale commonly used datasets. 
In contrast, for genius, arxiv-year, snap-patents, pokec, and wiki proposed in a recent work \cite{lim2021large}, the results imply that aggregated features are not enough signal for predicting node labels. 
Another example is GloGNN++ undirected which achieves the best results on genius and pokec but struggles in homomorphic graphs such as citeseer, coraML, and ogbn-arxiv.

These results validate that an appropriate selection of node representations and edge direction awareness is necessary to obtain state-of-the-art classification quality.
In summary, we observe that A2DUG stably perform well in various graphs, whereas no single existing method stably obtains state-of-the-art classification results.

\subsection{Q2. Scalability}
\label{ssec:performance_A2DUG}
Table \ref{tb:efficiency} shows the training time of existing methods and \textsc{A2DUG}. 

\smallskip \noindent \textbf{A2DUG scales well to the largest dataset wiki, which has 0.3 billion edges. }
In small-scale and middle-scale datasets, A2DUG runs within comparable time with existing methods in terms of both time per epoch and total time. 
Since A2DUG utilizes only MLPs in the model architecture\footnote{The time per epoch correlates with the number of MLP layers, which we adjust as a hyperparameter to optimize node classification quality. However, we observe that this adjustment does not markedly impact the training time.}, it does not require a significant increase in training time compared with other methods. 
Also, it can adopt minibatch training and thus scales well to large-scale datasets. 
In large-scale datasets, \textsc{A2DUG} needs longer training time per epoch than other methods because it combines both aggregated features and adjacency lists in directed/undirected graphs\footnote{In scenarios where the graph size is substantial, there is a modest increase in the time required to transfer matrices from the CPU memory to the GPU memory for each minibatch. This aspect slightly extends the time per epoch for A2DUG, which is attributed to its utilization of a larger set of node representations, enhancing the method's comprehensiveness.}.

The GNNs for directed graphs DGCN, DiGraph, and DiGraphIB do not scale to middle-scale graphs.
DGCN \cite{tong2020directed} calculates the second-order proximity matrices, which requires unacceptable computational cost $O(n^2)$. DiGraph and DiGraphIB \cite{tong2020digraph} calculate the eigenvalue decomposition of an adjacency matrix, which requires $O(n^3)$. These procedures are impractical even for middle-scale graphs.

In summary, A2DUG demonstrates strong performance on diverse datasets without imposing a significant time overhead. 

\begin{table*}[t]
    \caption{Comparison of the original A2DUG and its extension with learnable layer selection parameters. 
    }
    \centering
\begin{tabular}{l|ccc|ccc}
\toprule
 & genius & arxiv-year & ogbn-arxiv & snap-patents & pokec & wiki \\
\midrule
\textbf{A2DUG} & $89.85_{\pm3.15}$ & $59.14_{\pm0.48}$ & $69.51_{\pm0.24}$ & $72.38_{\pm0.10}$ & $82.48_{\pm0.10}$ & $65.13_{\pm0.07}$ \\
w layer selection & $90.22_{\pm0.32}$ & $60.18_{\pm0.16}$ & $69.52_{\pm0.22}$ & $72.20_{\pm0.06}$ & $82.67_{\pm0.11}$ & $64.80_{\pm0.08}$ \\
\bottomrule
\end{tabular}
\label{tb:layer_selection}
\end{table*}

\subsection{Q3. Ablation Study}
\label{ssec:ablation}
To investigate the performance contributions of aggregated features and adjacency lists in directed/undirected graphs, we conduct ablation studies with the variants removing directed graphs (``wo directed''), undirected graphs (``wo undirected''), aggregated features (``wo aggregation''), adjacency lists (``wo adjacency''), and transposed directed graphs (``wo transpose''). 
Table \ref{tb:ablation} shows the node classification results of the variants compared with \textsc{A2DUG}. 

\smallskip \noindent \textbf{Several datasets require all the combinations to achieve the state-of-the-art results. }
In cornell, squirrel, snap-patents, and wiki, the original \textsc{A2DUG} achieves the best classification results. 
This validates our assumption that aggregated features and adjacency lists in directed/undirected graphs complement each other.

\smallskip \noindent \textbf{The importance of aggregated features and adjacency lists depends on the characteristics of datasets. }
The variant ``wo aggregation'' obtains better results than ``wo adjacency'' in pokec and wiki. 
This means that adjacency lists play more important roles than aggregated features in these datasets. 
In contrast, the accuracy of ``wo adjacency'' is better than that of ``wo aggregation'' in snap-patent. 
This indicates that the importance of aggregated features and adjacency lists also depends on datasets. 
These observations validate that the adaptive effect control on aggregated features and adjacency lists is required. 

\smallskip \noindent \textbf{The importance of edge directions depends on the characteristics of datasets. }
In citeseer, coraML, genius, ogbn-arxiv, and pokec, the variant ``wo directed'' achieves the best results.
This implies that the information from undirected graphs is important for predicting the labels in these datasets. 
Also, directed graphs even work as noise in these datasets because the results of ``wo directed'' are slightly better than the original.
In contrast, in arxiv-year, the variant ``wo undirected'' achieves the best result. 
Even though the graph structure and node features of ogbn-arxiv are exactly the same as arxiv-year, interestingly, the importance of directed and undirected graphs is different due to the difference in their prediction targets, i.e., research fields and publication years, respectively. 
These observations clarify that the adaptive effect control on directed and undirected graphs is effective.

\smallskip \noindent \textbf{The transposed edges give an additional signal to classification. }
In most datasets, we observe that ``wo transpose'' obtains lower results than A2DUG.
This validates the effectiveness of using inverse edges in directed graphs.
Taking arxiv-year as an example, we conjecture that it is not only important which papers the paper cites but also which papers it is cited in, to correctly predict the publication year.

Consequently, all the combinations of node representations and edge direction are necessary to stably achieve the state-of-the-art classification performance across diverse datasets.

\subsection{Q4. Investigation on Model Design}
\label{ssec:model_validity}
While we simply concatenate all the combinations of node representations as shown in Eq. \eqref{eq:prediction}, it is a natural extension to introduce layer selection parameters that are expected to work as a soft selection of which node representation contributes to classification performance.
To validate the model design of A2DUG, we also develop and compare the extension that is formulated below:
\begin{align}
\label{eq:attention}
    \bmY = \text{MLP}_\text{final}\Bigl(\sigma(\alpha_0\bmH_{\bmX}\|\alpha_1\bmH_{\bmA}\|\alpha_2\bmH_{\bmA^{\top}}\|\alpha_3\bmH_{\bmA^\text{und}}\nonumber\\ 
    \|\alpha_4\bmH_\text{GNN}\|\alpha_5\bmH_{\text{GNN}^{\top}}\|\alpha_6\bmH_{\text{GNN}^\text{und}})\Bigr),
\end{align}
where $\alpha_0, \dots, \alpha_6$ are learnable layer selection parameters controlling the effects from node representations, which satisfy $\sum_{i=0}^6\alpha_i=1$. 

For this experiment, we use the same datasets and settings as Section \ref{ssec:effectiveness} and \ref{ssec:performance_A2DUG} and tune the hyperparameters of the extension. 
Table \ref{tb:layer_selection} shows the results comparing A2DUG and the extension (``w layer selection'').
The table demonstrates that there is no significant difference between A2DUG and the extension. 
This is because the final MLP layers of A2DUG can adaptively control the effects from node representations without the layer selection parameters as shown in Section \ref{ssec:performance_A2DUG}. 
Since the layer selection parameters do not offer benefits but complicate the model architecture, i.e., increase the number of learnable parameters, we use the simplest formulation shown in Eq. \eqref{eq:prediction} as our proposal A2DUG in this paper.
We leave the further exploration for the optimal combinations onto our future work.

\section{Discussion}
\label{sec:discussion}
We summarize and discuss the insights from the experiments. 

\smallskip \noindent \textbf{A2DUG presents a sensible choice if users are not familiar with the dataset characteristics. } 
This is because A2DUG achieves high-quality results across all datasets, as shown in Table \ref{tb:accuracy}. 
In practical settings, its stable performance would be highly appreciated, particularly when users lack knowledge about the specific dataset characteristics. 
We believe that A2DUG can reduce the burden of the manual model selection for users. 

\smallskip \noindent \textbf{Undirected edges play a significant role in classification within homophilous graphs. }
In our experiments, we used commonly used homophilic datasets, citeseer, coraML, and ogbn-arxiv. 
In Table \ref{tb:accuracy}, FSGNN which uses aggregated features in undirected graphs achieves the best accuracy. Also, the table shows that the variation ``wo undirected'' performs the worst. 
Remember that nodes in the same class tend to be connected in homophilous graphs. 
Neglecting the edge direction proves to be a suitable strategy for attaining high accuracy.

\smallskip \noindent \textbf{Depending on the prediction targets, non-homophilous graphs show various trends.}
First, we discuss the datasets in which the prediction target is related to time, i.e., arxiv-year and snap-patents (see Table~\ref{tb:04_dataset}). 
Since the edge direction indicates citations from older papers/patents to newer ones in the datasets, the edge direction plays a crucial role in predicting the correct labels.
Indeed, Table~\ref{tb:ablation} shows that accuracy significantly drops for the variant ``wo directed'' in arxiv-year and snap-patents. 

Second, in genius and pokec, we observe that the edge direction does not contribute to the classification quality since the variant ``wo directed'' achieves the highest results in Table \ref{tb:ablation}.
This indicates that the presence of edges serves as a signal for predicting the labels, rather than the direction of the edges, which is a different trend from arxiv-year and snap-patents. 

Through the above observations, we conclude that the importance of aggregated features and adjacency lists in directed/undirected graphs highly depends on datasets and their prediction targets.
As a result, it is important to adaptively control the effects from aggregated features and adjacency lists in directed/undirected graphs. 
Thus, \textsc{A2DUG} is robust across various prediction targets and achieves superior results to existing methods on datasets in which the combination of multiple factors is necessary since it can leverage the information from all of them.

\section{Conclusion}
\label{sec:conclusion}
We demonstrated that no existing methods stably obtain state-of-the-art results on various graphs since the importance of combinations depends on datasets. 
We proposed a simple yet scalable, effective, and robust method, \textsc{A2DUG}, that leverages all the combinations of node representations in directed/undirected graphs. 
Our empirical studies showed that A2DUG stably achieves state-of-the-art results across various datasets. 
Surprisingly, it outperforms state-of-the-art methods with large margins in arxiv-year, snap-patents, and wiki since it can adaptively control the effects stemming from aggregated features and adjacency lists in both directed/undirected graphs. 
Also, we observed that A2DUG scales well to graphs with $0.3$ billion edges. 
Finally, we discussed the insights of the dataset characteristics from our empirical studies.
We hope that our work motivates researchers to develop new GNN techniques beyond methods using either combination of node representations and edge directions.

\smallskip \noindent \textbf{Future work.}
Our study has several directions of future works.
First, we plan to extend A2DUG incorporates more sophisticated GNNs. Although this work showed the potential improvement of GNNs using all the combinations of node representations and edge directions by a simple method combining all of them, we do not focus on developing the optimal model architecture. Thus, we can expect to improve the accuracy and efficiency by developing new GNNs that fit our approach. 

Second, we employ neural architecture search in our method. We validated that the best architecture is different across datasets, for example in obgn-arxiv, directed edges are not effective. Therefore, it is expected to automatically select the best architecture and parameters without manual settings, which can improve accuracy.

Finally, our method can be extended to handle inequality and unfairness. In our and most existing empirical studies, ROC-AUC was used as the metric for the class-imbalanced dataset, genius. However, it is unsure how to select the metric and how to change the loss functions. 
Most GNNs minimize the cross-entropy loss between model outputs and training labels, which may result in the underrepresentation of minor classes during training. To address this, we plan to integrate loss functions tailored to specific metrics and datasets.

\section*{Acknowledgement}
This work was supported by JSPS KAKENHI Grant Number JP20H00583 and JST PRESTO JPMJPR18UD, Japan.

\bibliographystyle{IEEEtranS}
\bibliography{sample-base}

\end{document}